\documentclass{article}
\usepackage{ijcai26}

\usepackage[T1]{fontenc}
\usepackage[utf8]{inputenc}
\usepackage{textcomp}
\usepackage{multirow} 
\usepackage{placeins}
\usepackage{pifont}
\usepackage{array,tabularx,booktabs}
\usepackage{makecell} 
\usepackage{times}
\usepackage{soul}
\usepackage{url}
\usepackage[hidelinks]{hyperref}
\usepackage[utf8]{inputenc}
\usepackage[small]{caption}
\usepackage{graphicx}
\usepackage{amsmath}
\usepackage{amsthm}
\usepackage{booktabs}
\usepackage{algorithm}
\usepackage{algorithmic}
\usepackage[switch]{lineno}
\usepackage{bbding}
\usepackage{xcolor}
\usepackage{tikz}
\usetikzlibrary{calc}
\usepackage{wasysym}

\definecolor{green}{RGB}{0,100,0}
\definecolor{deepyellow}{RGB}{255,193,7}

\title{Federated Foundation Language Model
Post-Training Should Focus on Open Models}

\author{
Nikita Agrawal$^1$
\And
Ruben Mayer$^1$\\
\affiliations
$^1$University of Bayreuth\\
\emails
\{Nikita.Agrawal, Ruben.Mayer\}@uni-bayreuth.de
}

\begin{document}

\maketitle

\begin{abstract}

  Post-training of foundation language models has emerged as a promising research domain in federated learning (FL) with the goal to enable privacy-preserving model improvements and adaptations to users' downstream tasks. Recent advances in this area adopt centralized post-training approaches that build upon black-box and gray-box foundation language models, where there is limited access to model weights and architecture details. Although the use of such closed models has been successful in centralized post-training, their blind replication in FL raises several concerns. Our opinion is that using closed models in FL contradicts the core principles of federation, such as data privacy and autonomy. In this paper, we critically analyze the usage of closed models in federated post-training, and provide a detailed account of various aspects of openness and their implications for FL.

\end{abstract}

\section{Introduction}

Post-training of foundation language models has received increasing attention as models need to be adapted to downstream applications to ensure their effectiveness \cite{Tie2025ASO}. Traditional centralized machine learning approaches do not work well in privacy-sensitive settings, where data cannot be collected centrally and should remain private. Federated learning (FL) offers a promising alternative as a decentralized paradigm for training machine learning models on private data across multiple clients. Instead of exposing private data, clients only exchange gradients or model parameters~  \cite{mcmahan2023communicationefficientlearningdeepnetworks}. Combining FL with foundation models, federated post-training emerged, which involves adapting foundation language models within the FL framework for downstream tasks.

When deciding on what model and what methods to use for federated post-training, it is crucial to identify different types of foundation language models in terms of their \emph{openness}. In this paper, we classify them into four types according to their accessibility and licensing model: open-source, open-weight, gray-box, and black-box models. We find that the degree of openness of a model not only determines the methods that are available for post-training, but also impacts the autonomy of clients to use the post-trained model according to their needs (cf. Table~\ref{tab:modeltypes}).

\begin{table}[t]
\centering
\small
\caption{Analysis of model openness in federated post-training. \scriptsize\\\textit{\textbf{Notes:} 1. Open-source models allow inference logic to be customized. 2. API might provide access to logs. 3. Only possible if API exposes LoRA or adapter access. 4. Only possible if API allows it via managed fine-tuning. 5. Soft prompting possible if API allows.  6. Prompt Tuning only possible via textual prompting. FFT = Full Fine-tuning. RLHF = Reinforcement Learning from Human Feedback.}\\\textit{(Legend: {\color{green}\ding{51}}: Available ; {\color{deepyellow}\LEFTcircle}}: Partially Available ; {\color{red}\ding{55}}: Unavailable)}
\label{tab:modeltypes}

\begin{tabularx}{\columnwidth}{%
  >{\raggedright\arraybackslash}p{2.5cm} | 
  X | 
  X | 
  X | 
  X  
}
\toprule
 & \textbf{Open-Source} & \textbf{Open-Weight} & \textbf{Gray-Box} & \textbf{Black-Box} \\

\midrule

\textbf{I. Access} & & & &  \\

\small Weights & \makebox[\linewidth][c]{\color{green}\ding{51}}  & \makebox[\linewidth][c] {\color{green}\ding{51}}  &\makebox[\linewidth][c] {\color{deepyellow}\LEFTcircle}  & \makebox[\linewidth][c] {\color{red}\ding{55}}  \\

\small Architecture & \makebox[\linewidth][c] {\color{green}\ding{51}}  & \makebox[\linewidth][c] {\color{green}\ding{51}}  &\makebox[\linewidth][c] {\color{deepyellow}\LEFTcircle}  & \makebox[\linewidth][c] {\color{red}\ding{55}}  \\

\small Source Code & \makebox[\linewidth][c] {\color{green}\ding{51}} & \makebox[\linewidth][c] {\color{red}\ding{55}}  &\makebox[\linewidth][c] {\color{red}\ding{55}}  & \makebox[\linewidth][c] {\color{red}\ding{55}} \\

\small Training Data & \makebox[\linewidth][c] {\color{green}\ding{51}} & \makebox[\linewidth][c] {\color{red}\ding{55}} &\makebox[\linewidth][c] {\color{red}\ding{55}} & \makebox[\linewidth][c] {\color{red}\ding{55}} \\
\midrule

\textbf{II. Autonomy} & & & &  \\

\small On Premise & \makebox[\linewidth][c] {{\color{green}\ding{51}}\textsuperscript{1}} & \makebox[\linewidth][c] {{\color{green}\ding{51}}} & \makebox[\linewidth][c] {\color{red}\ding{55}} & \makebox[\linewidth][c] {\color{red}\ding{55}} \\

\small API inference & \makebox[\linewidth][c] {\color{green}\ding{51}} & \makebox[\linewidth][c] {\color{green}\ding{51}} & \makebox[\linewidth][c] {{\color{green}\ding{51}}\textsuperscript{2}} & \makebox[\linewidth][c] {\color{green}\ding{51}} \\

\midrule

\textbf{III. Post-Training} & & & &  \\

\small FFT & \makebox[\linewidth][c] {{\color{green}\ding{51}}} & \makebox[\linewidth][c] {{\color{green}\ding{51}}} & \makebox[\linewidth][c] {\color{red}\ding{55}} & \makebox[\linewidth][c] {\color{red}\ding{55}} \\

\small RLHF & \makebox[\linewidth][c] {{\color{green}\ding{51}}} & \makebox[\linewidth][c] {{\color{green}\ding{51}}} & \makebox[\linewidth][c] {\color{red}\ding{55}} & \makebox[\linewidth][c] {\color{red}\ding{55}} \\


\small LoRA  & \makebox[\linewidth][c] {{\color{green}\ding{51}}} & \makebox[\linewidth][c] {{\color{green}\ding{51}}} & \makebox[\linewidth][c] {{\color{green}\ding{51}}\textsuperscript{3}} & \makebox[\linewidth][c] {\color{red}\ding{55}} \\

\small Adapter Layers & \makebox[\linewidth][c] {{\color{green}\ding{51}}} & \makebox[\linewidth][c] {{\color{green}\ding{51}}} & \makebox[\linewidth][c] {{\color{green}\ding{51}}\textsuperscript{3}} & \makebox[\linewidth][c] {\color{red}\ding{55}} \\

\small Instruction Tuning & \makebox[\linewidth][c] {{\textcolor{green}{\ding{51}}}} & \makebox[\linewidth][c] {{\color{green}\ding{51}}} & \makebox[\linewidth][c] {{\color{green}\ding{51}}\textsuperscript{
4}} & \makebox[\linewidth][c] {{\color{red}\ding{55}}} \\

\small Prompt-Tuning & \makebox[\linewidth][c] {{\color{green}\ding{51}}} & \makebox[\linewidth][c] {{\color{green}\ding{51}}} & \makebox[\linewidth][c] {{\color{green}\ding{51}}\textsuperscript{5}} & \makebox[\linewidth][c] {{\color{green}\ding{51}}\textsuperscript{6}} \\


\midrule

\textbf{License} & \scriptsize Free & \scriptsize Restrictive
& \scriptsize Restrictive & \scriptsize Restrictive \\

\midrule

\textbf{Examples} & \scriptsize GPT-J, Mistral-7B & \scriptsize LlaMa family
& \scriptsize Together. AI & \scriptsize GPT-5 Claude\\

\bottomrule
\end{tabularx}
\vspace{-1.5em}

\end{table}

Recent centralized post-training approaches in frequently leverage black-box models \cite{yuksekgonul2025optimizing,khattab2023dspy}, where interaction is limited to an API that hides internal parameters (weights/architecture), making them inaccessible. This severely restricts the post-training methods. Yet, researchers deem black-box models acceptable and at the frontier of AI development, why they are heavily used in practice. Hence, researchers are now adopting black-box models for federated post-training, to leverage local data for post-training\cite{chen2025can,lin2023efficient}. However, this adaption requires a critical reflection on whether such approaches are suitable for the privacy-preserving use cases of FL.

\textbf{Our opinion is that federated foundation language model post-training should focus on open-source models.} Using black-box models makes clients rely on third-party services, introducing significant drawbacks such as potential data leakage, reduced transparency, and diminishing client autonomy that undermine the advantages of using FL. \emph{This is critical as it contradicts the fundamental goals and principles of FL.} Open-source models offer transparency, control, and alignment with the privacy-preserving principles of FL. They enable the use of robust privacy mechanisms, effective optimization techniques, and promote ethical AI development. This enhances the autonomy and sovereignty of clients over their data.

Our opinion is in line with similar statements from various groups in academia and industry that argue in favor of using open-source models \emph{in general}\cite{manchanda2025opensourceadvantagelarge,floridi2025open}. The particular focus of this paper is on the combination of \textbf{open-source vs. black-box models in federated post-training}, which is a question that requires more attention in the community.

 We make the following \textbf{contributions}. (1) We provide a detailed classification and analysis of the degree of openness of foundation language models, and their relation to federated post-training methods. (2) We critically discuss and analyze the implications of federated post-training of \emph{black-box} and \emph{gray-box} models. (3) We provide arguments for focusing on open models, showing that they support the core principles and motivations of federated learning better. (4) We perform a security and privacy analysis of federated post-training for different models and methods. (5) We provide a guide to select the most suitable type of foundation model (from open to closed) based on relevant decision criteria.

\section{Openness, Post-Training Methods and Federated Learning}

\subsection{Model Openness vis-à-vis Post-Training Methods}
\label{sec:FLinLLM}

\begin{figure}
     \centering
     \includegraphics[width=0.5\textwidth, trim=80pt 20pt 50pt 30pt, clip]{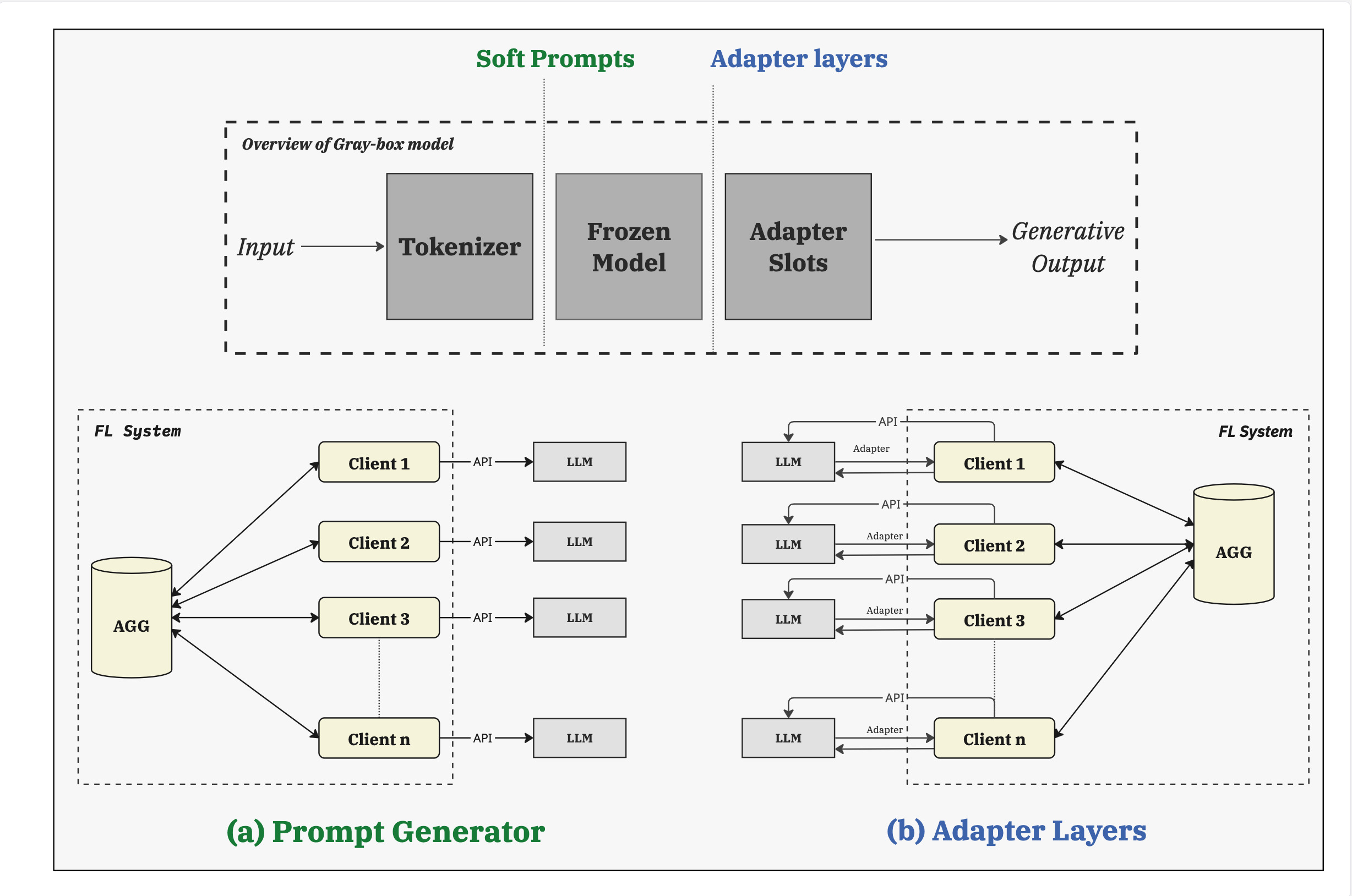}
     \vspace{-1em}
     \caption{Federated post-training of gray-box models. \textbf{Top panel}: Gray-box LLM, where the base model parameters are frozen and inaccessible, while post-training is enabled through APIs allowing soft prompts or lightweight adapter layers. \textbf{Bottom panel}: Two FL instantiations under the gray-box setting: (a) clients collaboratively learn soft prompts via local training and share only prompt models updates with a central aggregator, and (b) clients train local adapter layers attached to the frozen LLM and share only adapter parameters. In both cases, the base model remains frozen and proprietary, while the provider exposes only necessary trainable layers.}
     \label{fig:gray-box}
 \end{figure}

\paragraph{Degree of Openness}
 Open-source models \cite{aiAlliance2025aOpen,OSI-AI} provide full access to weights and source code, a well documented architecture, and training data under permissive licenses, allowing complete control over inference. Open-weight models \cite{open_weight_definition} also provide full access to weights and a well documented architecture but do not provide source code or any training data. Gray-box models provide partial access through APIs that allow for limited fine-tuning options, e.g., tuning of adapters. Figure~\ref{fig:gray-box} shows how restricted APIs in gray-box models can be leveraged in FL, where clients collaboratively train the prompt or adapter layers while the underlying foundation model remains frozen and is hosted centrally. Typically, gray-box models are actually open-weight models that are hosted by a service provider that offers managed model execution and post-training. Finally, black-box models are the most restrictive, allowing only the exchange of tokens without any access to model internals. They are fully hosted and managed by a service provider that does not offer an expressive API that goes beyond token exchange.

\paragraph{Post-Training Methods}
The accessibility and autonomy of the models define how they can be post-trained for downstream tasks.
Open-source and open-weight models can be post-trained using all possible methods, including full fine-tuning (FFT), parameter efficient fine-tuning (PEFT) such as Low-Rank Adapters (LoRA), adapter layers, prompt tuning (PT), instruction tuning (IT), and reinforcement learning from human-feedback (RLHF) \cite{rae2021scaling,zhao2023survey}. Gray-box models may provide token-level logits or an adapter interface, but they can only be accessed through an API that limits the fine-tuning options to what the provider exposes \cite{nakajima2024survey}. Like gray-box models, black-box models can also be accessed only via API; however, they do not expose any parameters, making it difficult to tune them beyond textual prompt tuning \cite{brown2020language,openai2024gpt4technicalreport}. 

\paragraph{Licensing}
Beyond access to source code and training data, another important distinction between open-source and open-weight models is their license. An open-source license must admit the unrestricted use, modification, and redistribution of the model \cite{OSI-AI}. This plays an important role for the post-training process itself, allowing for any needed modifications, but also for the further usage of the post-trained model. Any restrictions would impact the autonomy of users, even after post-training has been performed. E.g., when using Llama 3, one needs to follow the \emph{Meta Llama 3 Community License Agreement}. Under §2 (``Additional Commercial Terms''), this agreement states that ``\emph{[...] if [...] the monthly active users of the products or services made available [...] is greater than 700 million monthly active users [...], you must request a license from Meta [...] and you are not authorized to exercise any of the rights under this Agreement unless or until Meta otherwise expressly grants you such rights.}'' Hence, LLama 3 models fail to follow the open source principles---there are further restrictions in the license that we do not discuss in more detail here. In contrast, open-source models are released under open-source licenses that do not make such restrictions. An example is Mistral-7B that is released under the Apache-2.0 license. Beyond limitations of liability and warranty that protect the licensor, the license does not impose any restrictions on the use of the work or require any additional permissions under specific conditions.

A concise overview on accessibility, autonomy, and post-training techniques for each type of model (open-source, open-weight, gray-box, and black-box) can be found in Table \ref{tab:modeltypes}.

\subsection{Federated Learning on Foundation Models}
FL involves training a shared machine learning model across multiple distributed clients. Each client owns local training data that it does not share during training. In each training round, a server shares the global model with selected clients, which train it locally on their own data. After training, they send their model updates to the server, which combines them according to an aggregation strategy, e.g., Federated Averaging (FedAvg) \cite{mcmahan2023communicationefficientlearningdeepnetworks}, to improve the global model. Training is complete after multiple rounds when the model reaches an acceptable performance level. 

Post-training of foundation models on distributed datasets without collecting data centrally is essential in numerous real-world scenarios, particularly when handling sensitive information or complying with stringent data protection and governance regulations such as the GDPR \cite{gdpr2016} or the EU AI act \cite{euaiact2024}. 

The literature often distinguishes between \emph{cross-device} and \emph{cross-silo} FL \cite{10.1109/MCOM.005.2300467}. In cross-device settings, model training is performed on end-user devices, such as mobile phones, severely restricting feasible model sizes and training speed~\cite{10.1145/3652892.3700751}. In cross-silo FL, one assumes that clients have access to powerful machines or data centers, and the focus is on bridging data silos in a privacy-preserving manner. Such deployments are shown to be capable of training very large models with billions of parameters~\cite{sani2024photonfederatedllmpretraining}. As the focus of our paper is on post-training of very large foundation language models, we mainly consider cross-silo settings where computational capacities are typically high enough to run large models.

The implementation of FL in foundation language model post-training is a significant challenge given the large-scale and computational requirements of these models \cite{hu2024federated}. However, through the recent advent of strong small or medium-sized models and the increase in hardware capabilities, local post-training becomes a possibility. On the algorithmic side, PEFT techniques have become increasingly significant \cite{woiset2024,lin2023efficient}. They seek to reduce computational and communication overhead by training only a small subset of model parameters, while the majority remains frozen \cite{10.24963/ijcai.2025/1196}. 

The hardware requirements for post-training a foundation model in an FL setting largely depend on the size of the model, quantization precision, and the post-training method itself. For example, FFT on a 7-13B LLaMa style model using 16-bit precision needs tens of gigabytes of VRAM \cite{pan2403lisa} on a client node. In contrast, PEFT, especially 4-bit Quantized-LoRA (QLoRA) on a 7B LLaMa model, requires less than 12 GB of VRAM \cite{chai2023int2}. \cite{dettmers2023qlora} showed that a 4-bit NF4-quantized LLaMA model with LoRA adapters and optimizer states in bfloat16 can be post-trained using QLoRA with about 5 GB of GPU memory for the 7B parameter model, about 10 GB for the 13B model, 21 GB for the 33B model, and $<$ 48 GB for the 65B model on a single GPU. This shows that post-training even of larger foundation models is possible with modest hardware that can be found in cross-silo FL deployments. The hardware requirements of the server-node are not particularly vast as it does not train the model and is only involved in model aggregation and the orchestration of the training.

In addition to PEFT, personalized FL (PFL) represents a vital research domain with respect to foundation models \cite{lu2024zoopfl,wu2024fedbiot}. Recognizing the substantial variability in data distributions among diverse clients within a federated system (non-IID data), PFL aims to create models customized to the distinct requirements and data characteristics of individual clients. This typically entails a combination of global learning to acquire the fundamental knowledge within the entire federation and local adaptation to personalize the model for each client \cite{tan2022towards}. Recent variants of federated PEFT address this heterogeneity directly within the LoRA paradigm: SA-FedLoRA applies a simulated annealing schedule to dynamically adjust each client's parameter budget and mitigate client drift, while FlexLoRA allows clients to train local adapters at heterogeneous ranks, which are reconciled into a shared global adapter via SVD-based weight redistribution during aggregation \cite{bai2024federated,yang2025spdcflstepwiseparameterdropout}. 

\subsection{Federated Black-Box Post-Training Methods}

\begin{table}[t]
\centering
\small
\caption{Post-training methods for black-box foundation models in federated learning.}
\label{tab:black-box-foundation-FL}

\begin{tabularx}{\columnwidth}{
>{\raggedright\arraybackslash}X |
>{\raggedright\arraybackslash}X}
\toprule
\textbf{Post-Training Method} & \textbf{FL Papers} \\
\midrule

Continuous Soft Prompt Optimization &
FedBPT \cite{10.5555/3692070.3693989}, 
Fed-BBPT \cite{lin2023efficient}, 
ZooPFL \cite{lu2024zoopfl} \\
\midrule
Discrete Prompt Optimization &
FedOne \cite{wang2025fedonequeryefficientfederatedlearning}, 
FedDTPT \cite{wu2024feddtpt}, 
FedTextGrad \cite{chen2025can} \\
\midrule
Auxiliary Prompt Generate/Improvement Models &
 
\cite{lin2023efficient}, 
\cite{wu2024feddtpt}, 
\cite{chen2025can} \\

\bottomrule
\end{tabularx}

\vspace{-1.5em}
\end{table}

The reasons for the interest in black-box foundation language models are manifold \cite{matarazzo2025surveylargelanguagemodels,builtin_blackbox_ai}. Organizations that invest significantly in the development of these advanced models often claim and preserve ownership over their intellectual property and the technology itself. Facilitating access via APIs enables them to deliver their functionality as a service while concealing the intricate details of their models. Moreover, using these black-box models provides users with a convenient way to interact with advanced AI without requiring the substantial computing resources and expertise necessary to develop such extensive models independently. For that reason, we see a rapid increase of applications developed on these strong, yet non-transparent, foundation language models. With this growing application base, the question of post-training these black-box foundation models is receiving growing attention as can be seen in Table \ref{tab:black-box-foundation-FL}.

Recent research looks into how black-box foundation language models may be modified for the FL paradigm, given their popularity and success in centralized settings \cite{hu2024federated}. In FL, centralized post-training is usually adapted to the federated setting by training small auxiliary models to generate or improve prompts for the black-box model \cite{lin2023efficient,wu2024feddtpt,chen2025can}. This is because of the fundamental constraint that the internal parameters of a black-box model are inaccessible. 

FedBPT \cite{10.5555/3692070.3693989}, Fed-BBPT \cite{lin2023efficient}, Federated Discrete and Transferable Prompt Tuning (FedDTPT) \cite{wu2024feddtpt}, Federated Textual Gradient (FedTextGrad) \cite{chen2025can} and FedOne \cite{wang2025fedonequeryefficientfederatedlearning} aim to achieve efficient FL over black-box models that can be accessed only with APIs based on prompts. FedBPT and Fed-BBPT optimize continuous soft prompts, where clients estimate the gradients of the prompt parameters using a zeroth-order optimization. The local prompt updates are then aggregated on the server. \cite{lu2024zoopfl} investigate the effect of using zeroth-order optimization methods in personalized FL to work with black-box foundation language models. These methods approximate gradients by querying the model using inputs incorporated with an auto-encoder with low-dimensional and client-specific embeddings, enabling post-training without requiring direct access to the model’s internal parameters.

FedOne, FedDTPT, and FedTextGrad differ from FedBPT and Fed-BBPT since they operate on discrete prompts. Since they operate on discrete prompts, even zeroth-order optimization is not possible. In FedOne, clients must operate on search-based strategies such as block-wise exploration. Like FedBPT and Fed-BBPT, local prompt updates are aggregated to obtain a single prompt. In FedDTPT, a feedback loop mechanism that leverages the Masked Language Model API to guide prompt optimization is used during the client optimization phase. The server aggregates them using an attention method centered on semantic similarity to filter local prompt tokens, combined with embedding distance elbow detection and a DBSCAN clustering strategy to refine the filtering process. In FedTextGrad, clients make use of ``textual gradients'' for local prompt optimization. The server then aggregates the locally optimized prompts by summarizing them using the principle of uniform information density. 

\section{Open Models Fit Federated Learning Better}

\subsection{Alignment with Federated Learning Principles}

Although there are arguments for using black-box foundation language models (see Section~\ref{sec:counter-position}), we argue that the combination of black-box or gray-box foundation language models with FL is generally not a good fit. Relying on querying remote models hosted by a third party counteracts the benefits and motivations of using FL in the first place. 

\begin{table}[tp]
\caption{Effect of open-source criteria on federated learning principles when using the model for post-training. ``{\color{green}\ding{51}}'' means that the \textbf{violation} of this open-source criterion has severe impact on the corresponding federated learning principle; ``{\color{deepyellow}\LEFTcircle}'' means there is a limited or indirect effect; ``{\color{red}\ding{55}}'' means there is no effect.}
  \label{tab:openFL}
  
  \centering
  
  \begin{tabular}{p{2.4 cm}| p{1 cm}|p{1.4 cm}|p{1.8 cm}}

    \toprule
    \multirow{2}{*}{\makecell{Open-Source\\ Model Criteria}} & \multicolumn{3}{c}{\textbf{Federated Learning Principles}} \\
                     & Privacy & Autonomy & Heterogeneity \\
    \midrule
    Weights \& Code        &     \makebox[\linewidth][c]{\color{green}\ding{51}}     &    \makebox[\linewidth][c]{\color{green}\ding{51}}     &   \makebox[\linewidth][c]{\color{deepyellow}\LEFTcircle}                       \\ \hline
    License        &       \makebox[\linewidth][c]{\color{red}\ding{55}}   &     \makebox[\linewidth][c]{\color{green}\ding{51}}    &    \makebox[\linewidth][c]{\color{deepyellow}\LEFTcircle}                      \\ \hline
    Data          &   \makebox[\linewidth][c]{\color{red}\ding{55}}       &    \makebox[\linewidth][c]{\color{red}\ding{55}}      &     \makebox[\linewidth][c]{\color{red}\ding{55}}                      \\
   
    \bottomrule
  \end{tabular}

\end{table}

We analyze the impact of violated open-source criteria on three fundamental principles of FL, cf. Table~\ref{tab:openFL}. First, \emph{privacy}: One of the most important principles in FL is that clients keep their training data local and do not need to share it with any centralized training instance. Second, \emph{autonomy}: Clients decide themselves what training data to use and how to contribute to the joint training process. They also commonly own the trained model. Third, \emph{heterogeneity}: Clients with different kinds of data distribution can participate in the training, with the goal of covering a wide variety of private training data in the joint training process. 

\paragraph{Weights and code}
Without open weights, clients cannot directly access the foundation model for post-training. Instead, they must communicate with the model through the provider's API. This puts the client's privacy at risk as all communication is mediated by the API provider. The clients ability to perform fine-tuning tasks is constrained by the permissions and infrastructure of the API providers. This constrains the autonomy of the clients. Heterogeneity is not directly affected if model weights are not open: Post-training is still possible over diverse datasets. However, certain clients whose use-cases fall outside the provider's policies or are restricted by governing policies such as GDPR may be excluded from participation. The availability of model architecture and code is closely tied to the openness of the weights. Without this information, model usage and adaptation becomes significantly more difficult or even infeasible, forcing clients to rely on the model provider for fine-tuning.

\paragraph{Licenses}
Restrictive licensing does not impede the privacy of the client's post-training data. However, it does constrain their autonomy as they cannot freely decide how the fine-tuned models may be used. In other words, an open-weight model with a restrictive license can still be used for privacy-preserving federated post-training, but clients must carefully study the license and understand the impact of the restrictions on their intended use of the post-trained model. Licenses can provide a hurdle for certain participants to join the federated training, e.g. when they require a payment or when political conditions, such as international sanctions, restrict clients from obtaining them. This can reduce the variety of clients from participating in post-training and thus affect heterogeneity.

\paragraph{Data}
The openness of training data that is used for pre-training does not have any significant impact on federated post-training. Post-training is largely independent of pre-training: Clients can freely post-train the model with their private data regardless of whether the original pre-training data is open or not. It may generally be helpful to have information on the pre-training data, but this is not an issue that is specific to the federated setting.

\subsection{Closed Models Bring Many Risks}

\paragraph{Privacy and Security} FL enhances user privacy by keeping raw training data on local devices. Using black-box or gray-box models that are hosted by a third-party server puts the users' private data at risk, as the hosting party receives and processes plain-text input or tokens.  Hence, post-training on black-box models inherently leaks potentially sensitive information \cite{hanke2024}. One can try to add additional privacy-preserving mechanisms such as differential privacy \cite{behnia2022ew,zhao2023privacy} or rely on the service provider's guarantees to not look into the data, which can be supported by technical measures, e.g. via trusted computing \cite{FlowerIntelligence}. However, the fundamental privacy cornerstone of FL is that training data does not leave the client, and this is not possible with black-box and gray-box models. Beyond mere technical privacy, there are also regulatory difficulties when moving sensitive data, cf. privacy regulations such as GDPR. 
In terms of security, clients need to rely on mechanisms implemented by the black box model provider. Locally implemented security mechanisms in control of the clients \cite{ye2024emerging,Sun2025} are restricted to the components that are trained in a federated setting, i.e., mostly the prompt generator for black-box models or adapters for gray-box models.
It has to be noted that FL on open models is necessary, but not sufficient to ensure privacy, as various attacks are possible that require counter-measures. Then, however, clients are capable to apply and control defense mechanisms. We provide a more detailed analysis of privacy and security of federated post-training in Section~\ref{sec:security}.

\paragraph{Transparency and Explainability} Unlike open-source and open-weight foundation models, the limited access to black-box and gray-box models restricts understanding of their internal architecture and the reasons behind their output \cite{rane2024enhancing}. The lack of transparency hinders the systematic assessment and verification of the behavior and the analysis and mitigation of risks of these models according to predefined objectives. Relying on an external provider, clients do not even know for sure what system prompts were in place when using the models. 
Promising approaches to explain the output of a language model require access to the model parameters  \cite{ameisen2025circuit}.

\paragraph{Autonomy} One of the key advantages of FL is that clients retain control over their private data and key aspects of the training process. FL aims to promote the development of decentralized AI by returning data ownership to clients and reducing the dependency on a central authority \cite{flower_vana_2025,anelli2021put}. Outsourcing essential optimization decisions to a black-box model undermines this autonomy. If clients cannot retain complete control over the post-training process, one of the main motivations to use FL -- maintaining local client control -- is compromised. It raises the question whether the partial control that is left for the clients is sufficient to justify the use of FL in the first place.

\paragraph{Inefficient natural language interface} Communicating with black-box models is typically limited to natural language prompts. Arguably, natural language is not an effective way to program a computer system \cite{dijkstra997} due to its ambiguity, and the same argument holds for black-box foundation language models. Techniques that directly work on the model weights benefit from a more rigorous mathematical explanation of the optimization process.

\subsection{Open Models Have Advantages}

Our opinion is that black-box and gray-box foundation language models do not fit together with the core principles and goals of FL. In contrast, using open-source or open-weights models has many advantages and aligns very well with FL principles.

\paragraph{Local deployment provides control} If an open-source or open-weight model is deployed locally, we eliminate the need for API calls to black-box and gray-box models. Local deployment provides complete control over model training and data and allows for the use of additional privacy-preserving and security mechanisms. 
It should be noted that with more control, clients also have more responsibility. We discuss what this means for security and privacy in more detail in Section~\ref{sec:security}.

\begin{table}[t]
\centering
\small
\caption{Attack targets open-source, open-weights, gray-box, and black-box models, which \textbf{clients} can and need to protect during federated post-training.
\textit{({\color{green}\ding{51}}: Controllable (by clients) ; {\color{deepyellow}\LEFTcircle}: Partially controllable ; {\color{red}\ding{55}}: Out of control)}}
\label{tab:vulnerability}

\begin{tabularx}{\columnwidth}{%
  >{\raggedright\arraybackslash}p{2.5cm} | X | X | X | X }
\toprule
 \textbf{Targets} & \textbf{Open-Source} & \textbf{Open-Weight} & \textbf{Gray-Box} & \textbf{Black-Box} \\

\midrule

\small Base Model & \makebox[\linewidth][c]{\color{green}\ding{51}}  & \makebox[\linewidth][c] {\color{green}\ding{51}}  &\makebox[\linewidth][c] {\color{deepyellow}\LEFTcircle}  & \makebox[\linewidth][c] {\color{red}\ding{55}}  \\

\small Adapters & \makebox[\linewidth][c] {\color{green}\ding{51}}  & \makebox[\linewidth][c] {\color{green}\ding{51}}  &\makebox[\linewidth][c] {\color{deepyellow}\LEFTcircle}  & \makebox[\linewidth][c] {\color{red}\ding{55}}  \\

\small Prompt Generators & \makebox[\linewidth][c] {\color{green}\ding{51}} & \makebox[\linewidth][c] {\color{green}\ding{51}}  &\makebox[\linewidth][c] {\color{green}\ding{51}}  & \makebox[\linewidth][c] {\color{green}\ding{51}} \\

\bottomrule
\end{tabularx}
\vspace{-0.5em}
\label{tab:targets}
\end{table}

\paragraph{Broader range of post-training methods}  Beyond prompt tuning, open models support the full range of post-training techniques such as adapter layers and LoRA. Using PEFT, minimal parameter updates yield large task-specific performance improvements  \cite{hu2022lora,poth2023adapters}. PEFT techniques such as adapter layers and LoRA have been successfully extended to FL \cite{10.24963/ijcai.2025/1196,cai2023fedadapterefficientfederatedlearning}.

In terms of prompt tuning itself, model access facilitates highly effective techniques that work on soft prompts, such as prefix tuning \cite{li2021prefixtuningoptimizingcontinuousprompts} and P-Tuning \cite{DBLP:journals/corr/abs-2103-10385}.  \cite{lester2021power} indicate that prompt tuning, especially soft prompting, scales well with model size, achieving performance comparable to FFT. It has also motivated frameworks such as FedPrompt \cite{10095356} and PromptFL \cite{guo2023promptfl} that assume a frozen LLM with access to local gradients on each client and optimize prompt parameters to improve communication and computational efficiency. FL strategies such as FedSP  \cite{dong2023tunablesoftpromptsmessengers} that facilitate information exchange between clients and the aggregation server through soft prompts show the effectiveness of soft prompts in FL. Decomposed Prompt Tuning (DePT) combines soft prompting and LoRA shows great promise  \cite{shi2024deptdecomposedprompttuning}.



\subsection{Privacy and Security Considerations}
\label{sec:security}

First, we take a closer look at which parts of the model are in control of the clients. In Table~\ref{tab:targets}, we show that with a decreasing level of openness, clients have less control over the model. This means that when using gray-box and black-box models, clients need to trust the provider to take care of protecting the inaccessible model parts against attacks. On the other hand, with open-source and open-weight models, there is a larger attack surface during federated training, which clients and aggregation server need to protect against. Decentralizing the model training, hence, has two faces: It puts more control to the clients, but it may also increase the attack surface. Indeed, FL is known for its rich literature on privacy and security attacks and defenses, which we study in more detail in the following with regard to post-training methods.

To discuss attacks and defenses during FL post-training, we focus on the following attacks \cite{du2025privacy,hu2024federated,cheng2024towards}: 
1) \textit{Poisoning attacks:} manipulate the model to deteriorate downstream performance or install a backdoor, 2) \textit{membership inference:} infer whether a specific training sample was included in the training process, and 3) \textit{gradient inversion:} reconstruct original training data. To counter these attacks, various defenses have been proposed. We focus on the following defenses: 1) \textit{Differential privacy (DP):} injecting noise to hide individual data records, 2) \textit{Secure Aggregation:} privacy-preserving computation of an aggregated model without revealing individual clients' contributions, based on secure multi-party computation \cite{li2021secure}, 3) \textit{Homomorphic Encryption:} aggregation computations are performed directly on encrypted data, 4) \textit{Byzantine Aggregation:} robust aggregation protocols that tolerate Byzantine client behavior.
We analyze per post training method which of these attacks are possible and what measures clients can take to mitigate them \cite{cheng2024towards}.

\begin{table}
\def\check{\makebox[\linewidth][c]{\color{green}\ding{51}}}         
\def\parts{\makebox[\linewidth][c]{\color{deepyellow}\LEFTcircle}}  
\def\cross{\makebox[\linewidth][c]{\color{red}\ding{55}}}           
\centering
\small
\caption{Privacy and security analysis of federated post-training with respect to the fine-tuning method. We analyze which attacks \textbf{clients} can and need to protect against, and which methods could be effective in doing so. \scriptsize\\\\\textit{Legend: ({\color{green}\ding{51}}: Yes; {\color{deepyellow}\LEFTcircle}: Partially; {\color{red}\ding{55}}: No)}}
\label{tab:security}

\begin{tabularx}{\linewidth}{%
  >{\raggedright\arraybackslash}p{1.45cm} | X | X | X | X | X | X }

\toprule
& \textbf{FFT} &  \textbf{LoRA} & \textbf{Adap.} & \textbf{IT} &  \textbf{RLHF} & \textbf{PT} \\
\midrule

\textbf{Attacks} \\

\small Poisoning   &    \check & \check & \check & \check & \check & \check \\
\small Mem. Inf.   &    \check & \check & \check & \parts & \parts & \parts \\
\small Grad. Inv.  &    \check & \check & \check & \parts & \parts & \parts \\

\midrule

\textbf{Defenses} \\

\small DP               & \check & \check & \check & \parts & \parts & \parts \\
\small Sec. Agg.    & \check & \check & \check & \parts & \parts & \parts \\
\small HE               & \check & \check & \check & \parts & \parts & \parts \\
\small Byz. Agg.   & \check & \check & \check & \check & \check & \check \\

\bottomrule
\end{tabularx}
\vspace{-0.5em}
\end{table}

Table~\ref{tab:security} provides an overview of our results. As can be seen, the attacks to be protected against and the defenses available to the clients decrease with the degree of control over the model. While FFT and PEFT methods such as LoRA and Adapter layers have the largest attack surface, clients also have access to the full arsenal of protective measures. With less control over the model, attack surfaces and available protections also become more constrained. Instruction Tuning, RLHF and Prompt Tuning may or may not contain private or otherwise sensitive data; hence corresponding privacy attacks and defenses may or may not need to be considered. 

\section{Black-Box Foundation Language Models Have Many Advantages }
\label{sec:counter-position}

As can be seen in Table \ref{tab:black-box-foundation-FL}, black-box models remain widely used despite the risk they carry. Their adoption is driven by several advantages, such as strong performance, ease of deployment, and readily available serving stack.

Black-box foundation models often demonstrate superior performance compared to open-source alternatives. However, this is not always the case. Many open source models such as DeepSeek \cite{Guo_2025} consistently come extremely  close to black-box models.  Open-source models show strong performance in few-shot and zero-shot learning scenarios, allowing rapid adaptation to a wide range of downstream tasks.

Another important advantage of black-box models is their accessibility. Using hosted APIs removes the need for users to deploy expensive hardware infrastructure or maintain complex serving stacks locally. Service providers manage model updates, deployment, and operational maintenance, enabling users to access state-of-the-art models with relatively low setup effort. Furthermore, modern prompt engineering and in-context learning approaches allow users to adapt these systems without directly modifying model parameters. However, the practical benefits of hosted models depend strongly on the deployment setting. Recent advances in LLM serving stacks such as NVIDIA Triton \cite{triton_inference_server} and vLLM \cite{kwon2023efficient} make it easy for open-source models to be hosted privately. In addition, the terms of use of black-box model providers may restrict access, e.g., due to trade laws and sanctions \cite{openai2024gpt4technicalreport}.

Black-box model providers often supervise their deployment and maintenance in order to ensure security, alignment, and legal compliance, by observing ethical and legal guidelines \cite{hanke2024,laskar2023realWorld}. However, these safeguards can easily be breached with simple prompting attacks \cite{lapid2024opensesameuniversalblack,chao_jailbreaking_2024}. Users must ultimately trust the provider’s guarantees regarding data handling and privacy preservation, as the underlying model and infrastructure remain inaccessible. In contrast, locally deployed open models provide users with stronger control over privacy, security, and compliance mechanisms, which is particularly relevant in federated learning settings \cite{hanke2024,openai2024gpt4technicalreport}.
\section{A Guide to Model Selection }

\begin{figure}[!th]
\vspace{-0.2em}
\centering
\includegraphics[width=0.5\textwidth, trim=6pt 28pt 2pt 6pt, clip]{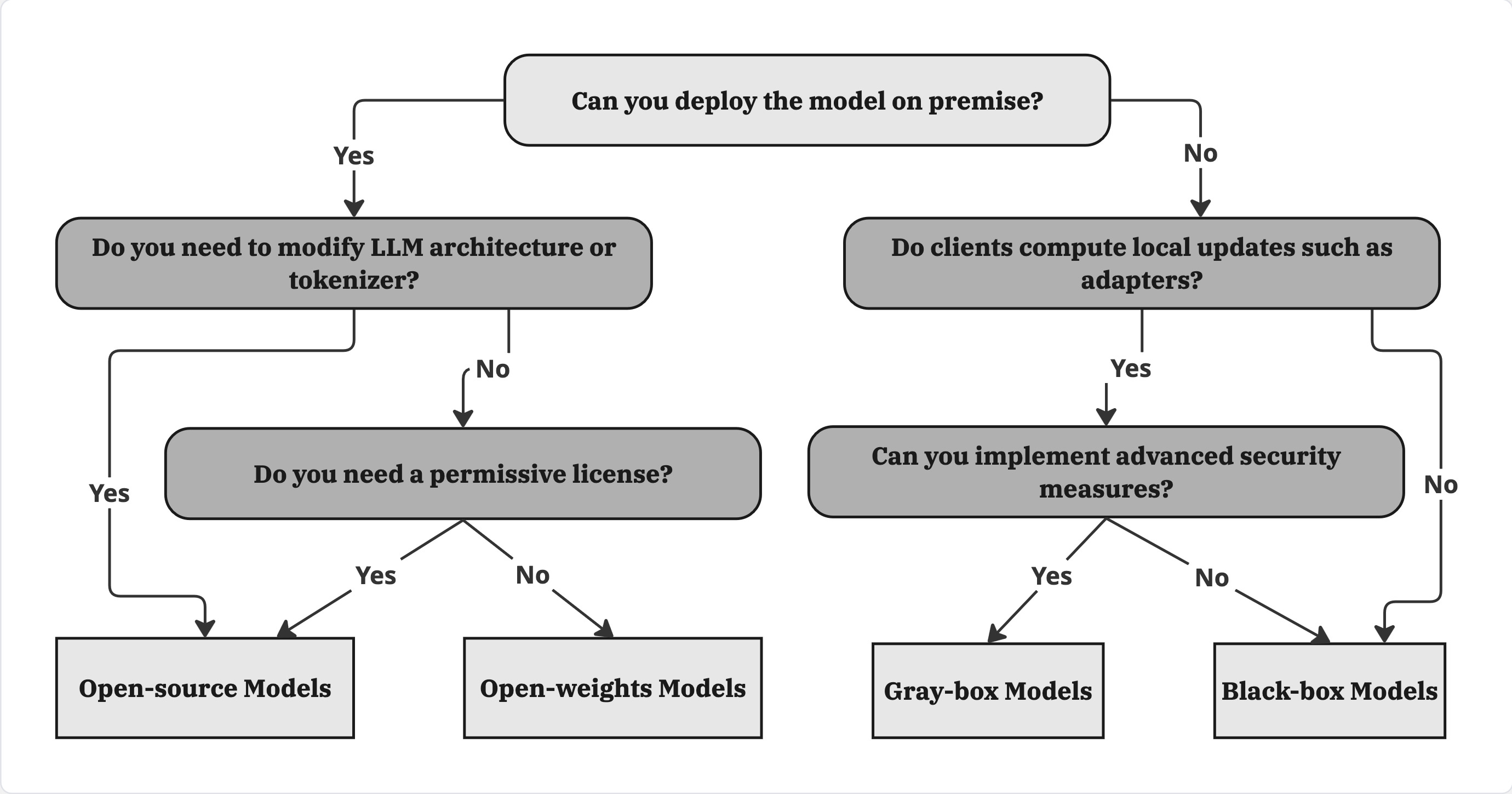}
\vspace{-1.2em}
\caption{Decision tree for selecting foundation model type based on privacy, security, autonomy, and heterogeneity}
\label{decision-tree:model}
\vspace{-1.0em}
\end{figure}

Following the discussions in this paper, users interested in federated foundation language model post-training may ask: \emph{Which type of model (open or closed) should I use in my specific setting?} The answer is multifaceted: Local resources, target model size, customization needs, licensing requirements, and the willingness to take responsibility for security and privacy. As a practical guide, we distill the discussion into a decision tree (Figure~\ref{decision-tree:model}) that aims to provide a high-level overview of important questions to be considered and point to the possibilities of open-source, open-weights, gray-box or black-box models. Together with Table~\ref{tab:modeltypes}, it can be used to make a first plan on what kind of model to combine with what kind of post-training method. The decision tree favors open models as the default option and only resorts to closed models if absolutely necessary. This reflects our stand that open models suit federated learning principles better.

\section{Conclusion}

In this paper, we discuss the following question: Should federated foundation language model post-training adopt black-box approaches? Our stand is clear: \emph{``No, it shouldn't.''} We provide a comprehensive analysis that illustrates how black-box models undermine the fundamental goals of FL, especially privacy and autonomy. Based on a detailed analysis of aspects of openness in foundation models, we argue which of them are critical to which goal of FL. This allows for a more fine-grained assessment of models that offer \emph{partial} openness (e.g., open weights but restricted licenses). This way, our work provides a tool-box that helps researchers and practitioners to make an informed decision on which foundation models to consider for their work on federated post-training.

Our analysis primarily focuses on the deployment and post-training of standalone foundation language models. However, modern large-scale AI systems increasingly operate as compound systems consisting of multiple interacting components, including auxiliary draft models, routing mechanisms, safety and alignment layers, retrieval systems, hardware-specific inference kernels, and external tool integrations. In such settings, the degree of openness may differ across individual system components, making the distinction between open and closed models less clear-cut than discussed in this paper. Extending federated post-training analyses toward these multi-component AI systems represents an important direction for future work.


\bibliographystyle{named}
\bibliography{bibliography}

\end{document}